\documentclass[runningheads]{llncs}
\usepackage[T1]{fontenc}
\usepackage{graphicx}
\usepackage{amsmath}
\usepackage{float}

\usepackage{ulem}
\usepackage{wrapfig}
\usepackage{amssymb}
\usepackage{booktabs}
\usepackage{makecell}
\usepackage{subcaption}
\usepackage[table]{xcolor}
\usepackage{colortbl}
\usepackage{utfsym}

\newcommand\net{KGA-Net}

\usepackage{pifont} 
\newcommand{\bd}[1]{\textbf{#1}}
\newcommand{\app}{\raise.17ex\hbox{$\scriptstyle\sim$}}

\makeatletter
\DeclareRobustCommand\onedot{\futurelet\@let@token\@onedot}
\def\@onedot{\ifx\@let@token.\else.\null\fi\xspace}

\makeatother

\begin{document}
\title{Boosting Breast Ultrasound Video Classification by the Guidance of Keyframe Feature Centers}
\titlerunning{{\net} for Breast Ultrasound Video Classification}
%
\author{Anlan Sun\thanks{Equal contribution.}\inst{1} \and
Zhao Zhang$^\star$\inst{2} \and
Meng Lei\inst{2} \and
Yuting Dai\inst{1} \and
Dong Wang\inst{3} \and
Liwei Wang\inst{2, 3}}
\authorrunning{Anlan et al.}
%
\institute{
Yizhun Medical AI Co., Ltd \\
\email{\{anlan.sun, yuting.dai\}@yizhun-ai.com} \and
Center for Data Science, Peking University\\
\email{\{zhangzh, leimeng\}@stu.pku.edu.cn} \and
National Key Laboratory of General Artificial Intelligence,
School of Intelligence Science and Technology, Peking University\\
\email{\{wangdongcis, wanglw\}@pku.edu.cn}
}
\maketitle              
\begin{abstract}

Breast ultrasound videos contain richer information than ultrasound images, therefore it is more meaningful to develop video models for this diagnosis task. However, the collection of ultrasound video datasets is much harder. In this paper, we explore the feasibility of enhancing the performance of ultrasound video classification using the static image dataset. To this end, we propose {\net} and coherence loss. The {\net} adopts both video clips and static images to train the network. The coherence loss uses the feature centers generated by the static images to guide the frame attention in the video model. Our {\net} boosts the performance on the public BUSV dataset by a large margin. The visualization results of frame attention prove the explainability of our method. \textit{The codes and model weights of our method will be made publicly available.}

\end{abstract}

\section{Introduction}

Breast cancer is a life-threatening disease that has surpassed lung cancer as leading cancer in some countries and regions~\cite{statistic}. Breast ultrasound is the primary screening method for diagnosing breast cancer, and accurately distinguishing between malignant and benign breast lesions is crucial. This task is also an essential component of computer-aided diagnosis.
Since each frame in an ultrasound video can only capture a specific view of a lesion, it is essential to aggregate information from the entire video to perform accurate automatic lesion diagnosis. Therefore, in this study, we focus on the classification of breast ultrasound videos for detecting malignant and benign breast lesions.

Despite the fact that ultrasound videos contain more information than static images, most previous studies have focused on static image classification~\cite{ori_7,ori_9,ori_11}. One major difficulty in using ultrasound videos for diagnosis lies in the collection of video data with pathology gold standard results. Firstly, during general ultrasound examinations, sonographers usually only record keyframe images and not entire videos. Secondly, for prospectively collected videos, additional effort must be made to track the corresponding pathological results. As a result, while there are many breast ultrasound image datasets~\cite{busi,busis}, video datasets are scarce. Currently, there is only one breast video dataset~\cite{busv} available, which is relatively small, containing only 188 videos.

\begin{figure}[t]
\includegraphics[width=\textwidth]{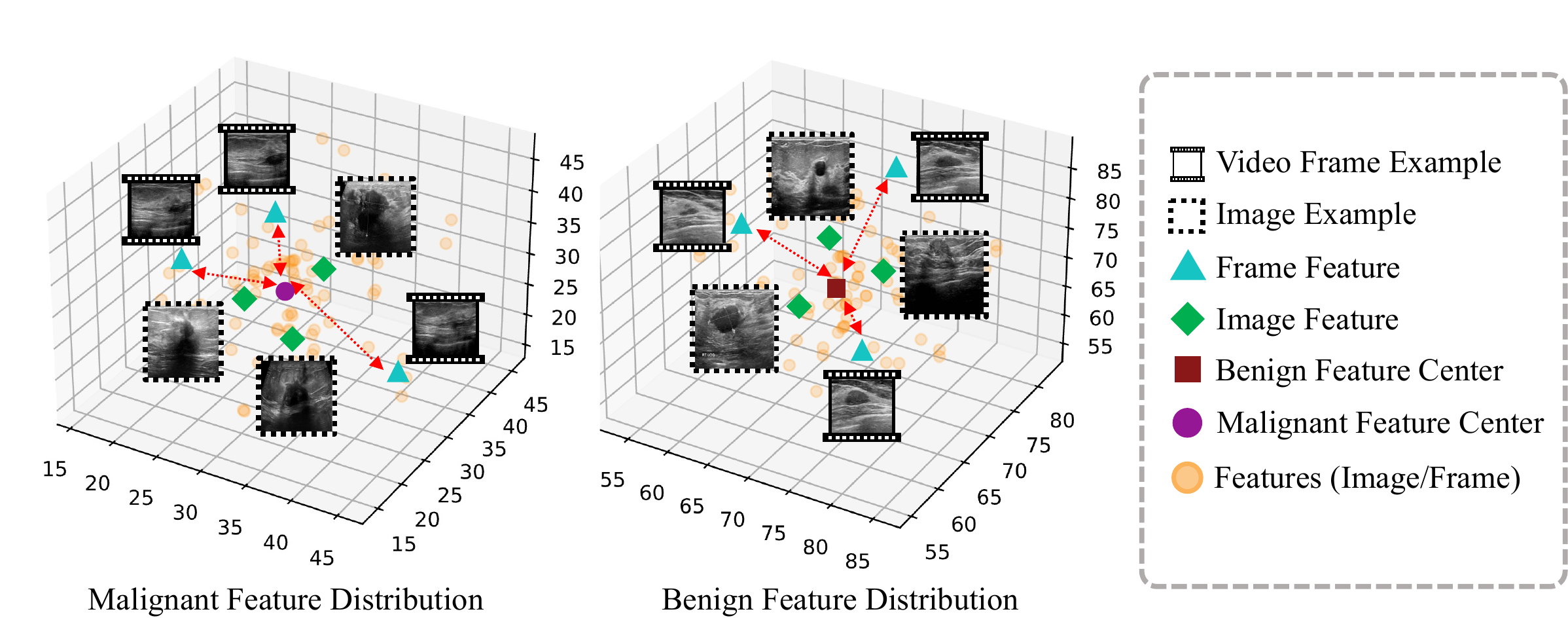}
\caption{Feature distribution of video frames from BUSV~\cite{busv} and static images from BUSI~\cite{busi}. We use a 2D ResNet trained on ultrasound images to get the features.}
\label{fig:feature}
\end{figure}

Given the difficulties in collecting ultrasound video data, we investigate the feasibility of enhancing the performance of ultrasound video classification using a static image dataset. To achieve this, we first analyze the relationship between ultrasound videos and images. The images in the ultrasound dataset are keyframes of a lesion that exhibit the clearest appearance and most typical symptoms, making them more discriminative for diagnosis. Although ultrasound videos provide more information, the abundance of frames may introduce redundancy or vagueness that could disrupt classification. 
From the aspect of feature distribution, as shown in Fig.~\ref{fig:feature}, the feature points of static images are more concentrated, while the feature of video frames sometimes are away from the class centers. Frames far from the centers are harder to classify.
Therefore, it is a promising approach to guide the video model to pay more attention to important frames close to the class center with the assistance of static keyframe images. Meanwhile, our approach aligns with the diagnosis of ultrasound physicians, automatically evaluates the importance of frames, and diagnoses based on the information of key frames. Additionally, our method provides interpretability through key frames.

In this paper, we propose a novel Keyframe Guided Attention Network (\net) to boost ultrasound video classification. Our approach leverages both image (keyframes) and video datasets to train the network. To classify videos, we use frame attention to predict feature weights for all frames and aggregate them to make the final classification. The feature weights determine the contribution of each frame for the final diagnosis.
During training, we construct category feature centers for malignant and benign examples respectively using center loss~\cite{ori_14} on static image inputs and use the centers to guide the training of video frame attention.
Specifically, we propose coherence loss, which promotes the frames close to the centers to have high attention weights and decreases the weights for frames far from the centers. 
Due to the feature centers being generated by the larger scale image dataset, it provides more accurate and discriminative feature centers which can guide the video frame attention to focus on important frames, and finally leads to better video classification. 

Our experimental results on the public BUSV dataset~\cite{busv} show that our {\net} significantly outperforms other video classification models by using an external ultrasound image dataset. 
Additionally, we visualized attention values guided by the coherence loss. The frames with clear diagnostic characteristics are given higher attention values. This phenomenon makes our method more explainable and provides a new perspective for selecting keyframes from video.

In conclusion, our contributions are as follows:
\begin{enumerate}
\item We analyze the relationship between ultrasound video data and image data, and propose the coherence loss to use image feature centers to guide the training of frame attention.
\item We propose {\net}, which adopts a static image dataset to boost the performance of ultrasound video classification. {\net} significantly outperforms other video baselines on the BUSV dataset.
\item The qualitative analysis of the frame attention verifies the explainability of our method and provides a new perspective for selecting keyframes.
\end{enumerate}

\section{Related Works}

\noindent\bd{Breast Ultrasound Classification.} Breast ultrasound (BUS) plays an important supporting role in the diagnosis of breast-related diseases. Recent research demonstrated the potential of deep learning for breast lesion classification tasks~\cite{ori_4,ori_5,ori_10,ori_11,ori_12}. \cite{ori_4,ori_5} design ensemble methods to integrate the features of multiple models to obtain higher accuracy. ~\cite{ori_10,ori_11,ori_12} utilize multi-task learning to improve the model performance. However, all of them are based on image datasets, such as BUSI~\cite{busi}, while few works focus on the video modality. \cite{bus-video1} designed a pre-training model based on contrastive learning for ultrasound video classification. \cite{ori_13,bus-video2} develop a keyframe extraction model for ultrasound videos and utilized the extracted keyframes to perform various classification tasks. However, these methods rely on keyframe supervision, which limits their applicability. Fortunately, the recent publicly available dataset BUSV~\cite{busv} has made the research on the task of BUS video-based classification possible. In this paper, we build our model based on this dataset.

\noindent\bd{Video recognition based on neural networks.} Traditional methods are based on Two-stream networks~\cite{twostream,tsn,st_resnet}. Since I3D~\cite{i3d} was proposed, 3D CNNs have dominated video understanding for a long time. ~\cite{r2p1d,csn} decompose 3D convolution in different ways to reduce computation complexity without losing performance. ~\cite{slowfast} designed two branches to focus on temporal information and spatial features, respectively. However, 3D CNNs have a limited receptive field, and thus struggle to capture long-range dependency. Vision Transformers~\cite{vit,swin} have become popular due to their excellent capability of aggregating spatial-temporal information. In order to reduce computational complexity brought by global attention, MViT~\cite{mvit} used hierarchical structure by reducing spatial resolution and Video Swin~\cite{videoswin} introduced 3D shifted window attention. Our proposed KGA-Net is a simple framework that aggregates multi-frame features based on the frame attention module.

\section{Methodology}

\begin{figure}[t]
\includegraphics[width=\textwidth]{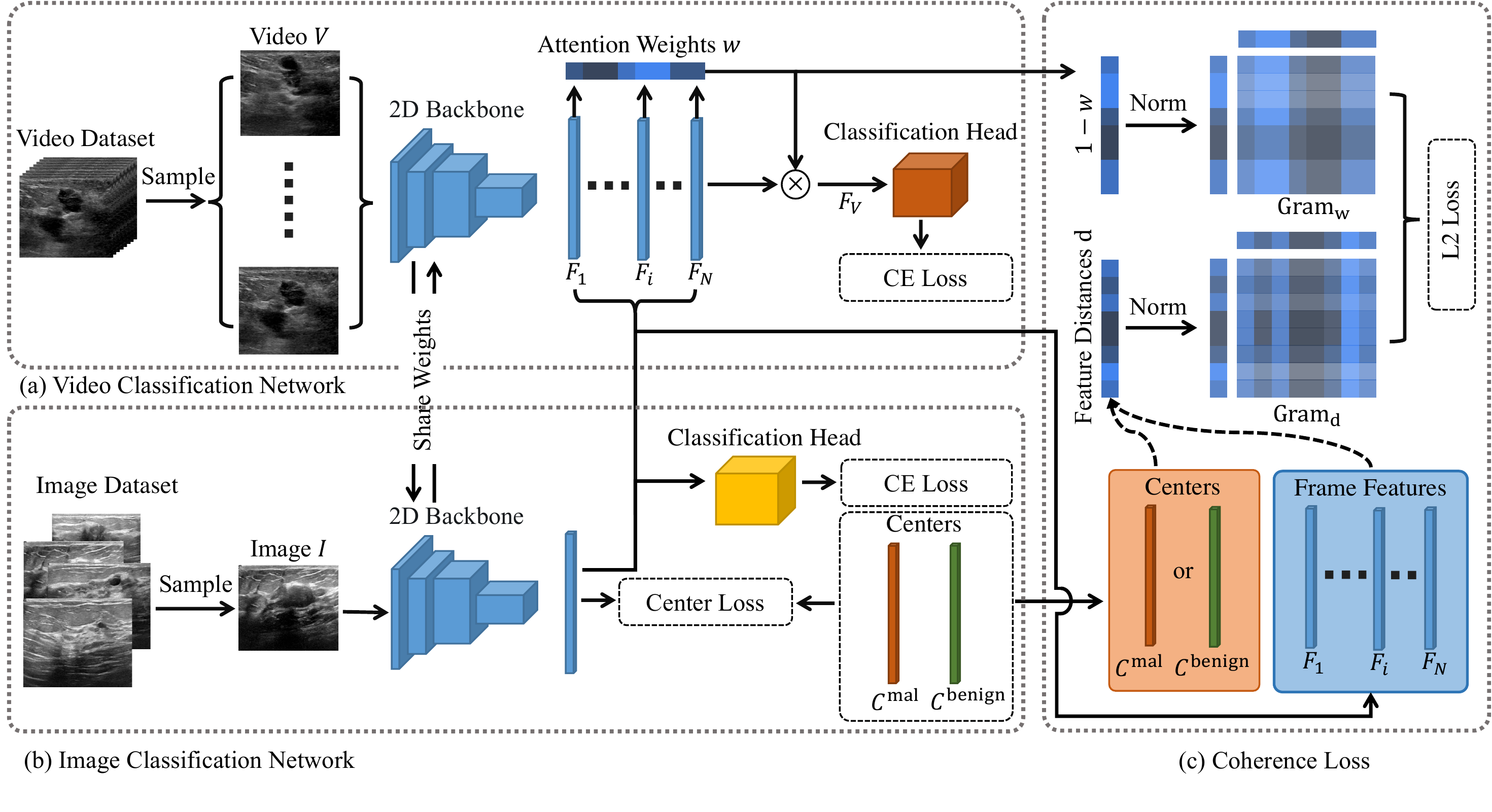}
\caption{Overview of our proposed keyframe-guided attention network.}
\label{fig:overview}
\end{figure}

As shown in Fig.~\ref{fig:overview}, our {\net} takes the video inputs and static image inputs simultaneously to train the network. The coherence loss is proposed to guide the frame attention by using the feature centers generated by the images.
We will then elaborate on each component in the following sections.

\subsection{Video and Image Classification Network}

\noindent\bd{The video classification network} is illustrated in Fig.~\ref{fig:overview} (a). The model is composed of a 2D CNN backbone, a frame attention module, and a classification head.
For an input video clip $V$ composed of $N$ frames, it is first processed by the backbone network and the feature vectors of the frames $\{ F_i \}_{i=1}^N$ are obtained. Then, the frame attention module predicts the attention weight for each frame using a FC and sigmoid layer, and then the features are aggregated by the weights to form an integrated feature vector. Formally, 
\begin{equation}
w_i = \text{Sigmoid} ( \text{FC} ( F_i ) )
\end{equation}
where $w_i$ denotes the weight for the $i_{\text{th}}$ frame and FC is the fully-connected layer. Then, the features are aggregated by $F_V = \sum_{i=1}^N {w_i \cdot F_i}$. Finally, the classification head is applied to the final result of lesion classification. To train the model, the cross-entropy loss (CE Loss) is applied to the classification prediction of the video.

\noindent\bd{The image classification network} is used to assist in training the video model. We use the same 2D CNN as the backbone network in the video classification network. The model weights are shared for the two backbones for better generalization. To promote the formation of feature centers, we apply the center loss~\cite{ori_14} to the image model besides the cross-entropy loss. In addition, the frame-level cross-entropy loss is also applied to the video frames to facilitate training.

\subsection{Training with Coherence Loss}

In this section, we introduce the coherence loss to guide the frame attention with the assistance of the category feature centers. We use the same method as center loss~\cite{ori_14} to obtain the feature centers for the malignancy and benign lesions, which are denoted as $\mathcal{C}^\text{mal}$ and $\mathcal{C}^\text{benign}$, respectively.

The distances of frame features and the feature centers can measure the quality of the frames. The frame features close to the centers are more discriminative for the classification task. Therefore, we use these distances to guide the generation of frame attention. Specifically, we push the frames close to the centers to have higher attention weights and decrease the weights far from the centers. To do this, for each video frame with feature $F_i$, we first calculate the feature distance from its corresponding class center. Formally, 
\begin{equation}
d_{i} = \| F_i - \mathcal{C}^Y \|_2,
\end{equation}
where $Y \in \{ \text{mal}, \text{benign} \}$ is the label of the video $V$ and $d_i$ is the computed distance of frame $i$. 

Afterward, we apply coherence loss to the attention weights $\textbf{w}=[ w_1, w_2, ..., w_N ]^\intercal $ to make them have a similar distribution with the feature distances $\textbf{d}=[ d_1, d_2, ..., d_N ]^\intercal $. To supervise the distribution, the coherence loss is defined as the L2 loss of the gram matrix of these two vectors
\begin{equation}
\text{L}_\text{Coh} = \| \text{Gram}_\textbf{w} - \text{Gram}_\textbf{d} \|_2,
\end{equation}
where
$\text{Gram}_\textbf{w} = \frac{(\mathbf{1}  - \textbf{w}) \cdot (\mathbf{1} - \textbf{w})^\intercal}{\| \mathbf{1} - \textbf{w} \|_2^2} $
is the gram matrix of normalized attention weights, and 
$\text{Gram}_\textbf{d} = \frac{\textbf{d} \cdot \textbf{d}^\intercal}{\| \textbf{d} \|_2^2 }$
is the gram matrix of normalized feature distances. Note that lower distances correspond to stronger attention, hence we use the opposite of $\textbf{w}$ to get $\text{Gram}_\textbf{w}$.

\subsection{Total Training Loss}

To summarize, the total training loss of our {\net}
\begin{equation}
L_\text{total} = L_\text{CE}^V + L_\text{CE}^I + L_\text{Center} + \lambda \cdot L_\text{Coh}.
\end{equation}
$L_\text{CE}^V$ and $L_\text{CE}^I$ denote the cross-entropy for video classification and image and frame classification. $L_\text{Center}$ means the center loss. $\lambda$ is the weight for coherence loss. Empirically, we set $\lambda=1$ in our experiments.

During inference, to perform classification on video data, the video classification network can be utilized individually for prediction.

\section{Experiments}

\subsection{Implementation Details}

\noindent\bd{Datasets.} We use the public BUSV dataset~\cite{busv} for video classification and the BUSI dataset~\cite{busi} as the image dataset. BUSV consists of 113 malignant videos and 75 benign videos. BUSI contains 445 images of benign lesions and 210 images of malignant lesions. For the BUSV dataset, we use the official data split in \cite{busv}. All images of the BUSI dataset are adopted to train our {\net}.

\noindent\bd{Model Details.} ResNet-50~\cite{he2016deep} pretrained on ImageNet~\cite{deng2009imagenet} is used as backbone. We use SGD optimizer with an initial learning rate of 0.005, which is reduced by 10$\times$ at the 4,000th and 6,000th iteration. The total learning iteration number is 8,000. The learning rate warmup is used in the first 1,000 iterations. 
For each batch, the video clips and static images are both sampled and sent to the network. We use a total batchsize of 16 and the sample probability of video clips and images is 1:1. We implement the model based on Pytorch and train it with NVIDIA Titan RTX GPU cards.

During inference, we use the video classification network individually. In order to satisfy the fixed video length requirement of MViT~\cite{mvit}, we sample up to 128 frames of each video to form a video clip and predict its classification result using all the models in experiments. 

\subsection{Comparison with Video Models}

In this section, we compare our {\net} with other competitive video classification models. Comparing with ultrasound-video-based work presents difficulty. \cite{bus-video1,bus-video2} is not accompanied by open-source code and relies on private datasets, making comparisons exceedingly challenging. \cite{ori_13} relies on a private dataset with keyframe annotations for supervised training. The released code does not include keyframe detection, which makes direct comparison impossible. Since the research on ultrasound video classification is uncomparable, we compare our method with other strong video baselines on natural images. The CNN-based models including I3D~\cite{i3d}, SlowFast~\cite{slowfast}, R(2+1)D~\cite{r2p1d} and CSN~\cite{csn} are involved. Meanwhile, the recently popular transformer-based model (MViT~\cite{mvit}) is also adopted. For a fair comparison, we use both the video and image data to train these models. The images are regarded as static videos to train the networks.
During evaluation, we report the metrics on the test set of BUSV. 

As shown in Table.~\ref{tab:sota}, by leveraging the guidance of the image dataset, our {\net} significantly surpasses all other models on all of the metrics. The video classification model of our {\net} is composed of a standard 2D ResNet-50 and a light feature attention module, while the baseline models are with net structures carefully designed for video analysis. Therefore, the success of our {\net} lies in the correct usage of the image guidance. The feature centers formed by the image dataset with larger data size and clear appearance effectively improve the accuracy of frame attention hence boosting the video classification performance.

\begin{table}[t]
\centering
\caption{\bd{Comparison with other video models.} Classification thresholds are determined by Youden index.}
\begin{tabular}{l|c|c|c|c}
\toprule
\makebox[0.15\textwidth][c]{Model} & \makebox[0.13\textwidth][c]{AUC(\%)} & \makebox[0.13\textwidth][c]{ACC(\%)} & \makebox[0.17\textwidth][c]{Sensitivity(\%)} & \makebox[0.17\textwidth][c]{Specificity(\%)} \\
\midrule
I3D~\cite{i3d} & 88.31 & 81.58 & 84.00 & 76.92\\
SlowFast~\cite{slowfast} & 82.54 & 79.49 & 76.92 & 84.62\\
R(2+1)D~\cite{r2p1d} & 86.46 & 81.58 & 84.00 & 76.92\\
CSN~\cite{csn} & 83.38 & 81.58 & 84.00 & 76.92\\
MViT~\cite{mvit} & 90.53 & 82.05 & 80.77 & 84.62\\
KGA-Net (Our) & {\textbf{94.67}} & {\textbf{89.74}} & {\textbf{88.46}} & {\textbf{92.31}}\\
\bottomrule
\end{tabular}
\label{tab:sota}
\end{table}

\subsection{Ablation Study}

In this section, we ablate the contribution of each key design in our {\net}. We observe their importance by removing these key components from the whole network. The results are shown in Table~\ref{tab:ablation}. The results of {\net} are shown in the last row in Table~\ref{tab:ablation}, while the components are ablated in the first three rows. We use the same training schedule for all of the experiments.

\noindent\bd{Image guidance} is the main purpose of our method. To portray the effect of using the image dataset, we train the {\net} using BUSV dataset alone in the first row of Table~\ref{tab:ablation}. Without the image dataset, we generate the feature centers from the video frames. As a result, the performance significantly drops due to the decrease in dataset scale. It also shows that the feature centers generated by the image dataset are more discriminative than that of the video dataset. It is not only because the lesion number of BUSI is larger than BUSV, but also because the images in BUSI are all the keyframes that contain typical characteristics of lesions.  

\noindent\bd{Frame attention and coherence loss} are two essential modules of our KGA-Net. We train a {\net} without the coherence loss in the third row of Table~\ref{tab:ablation}. In the second row, we further replace the feature attention module with feature averaging of video frames. It can be seen that both of these two modules contribute to the overall performance according to AUC and ACC. It is worth noting that these two models without coherence loss obtain very low sensitivity and high specificity, which means the model predictions are imbalanced and intend to make benign predictions. It is because that clear malignant appearances usually only exist in limited frames in a malignant video. Without our coherence loss or frame attention, it is difficult for the model to focus on typical frames that possess malignant features. This phenomenon certifies the effectiveness of our {\net} to prevent false negatives in diagnosis.

\begin{table}[t]
\centering
\caption{\bd{Ablation studies.} Model components are removed in the first three lines to analyze their contributions in {\net}. Classification thresholds are determined by Youden index.}
\begin{tabular}{l|c|c|c|c}
\toprule
Model &  AUC(\%) & ACC(\%) & Sensitivity(\%) & Specificity(\%)\\
\midrule
w/o image guidance              & 85.21 & 76.92 & 73.08 & 84.62\\
w/o coherence loss \& attention & 88.17 & 74.36 & 61.54 & \bd{100.0}\\
w/o coherence loss              & 92.90 & 87.18 & 80.77 & \bd{100.0}\\
\midrule
{\net} & \bd{94.67} & \bd{89.74} & \bd{88.46} & 92.31\\
\bottomrule
\end{tabular}
\label{tab:ablation}
\end{table}

\subsection{Visual Analysis}
\begin{figure}
\includegraphics[width=1.0\textwidth]{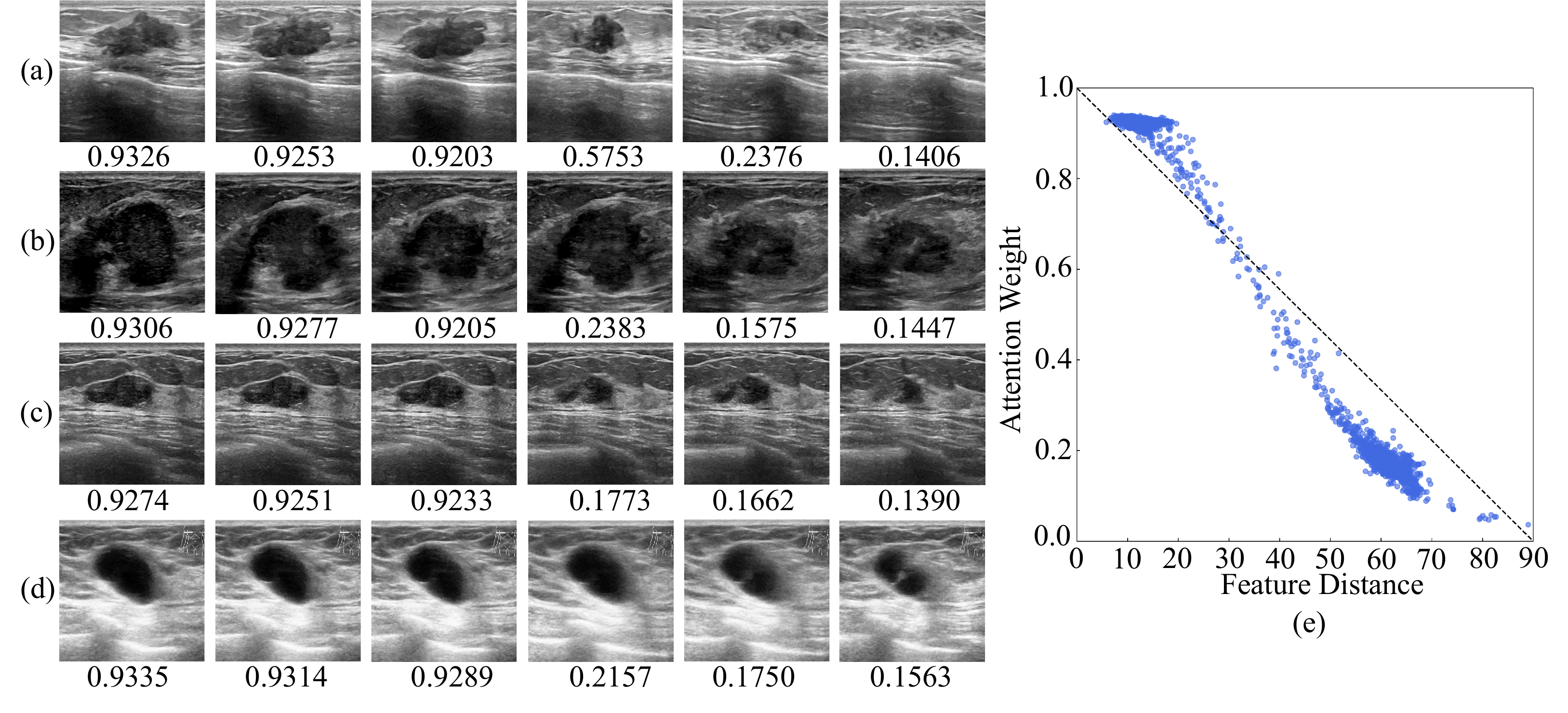}
\caption{\bd{Visual Analysis.} (a-d) Visualization of video frames and corresponding frame attention weights. (e) Relationship between attention weight and feature distance.}
\label{fig:vis}
\end{figure}

In Fig.~\ref{fig:vis}, we illustrate video frames with their corresponding frame attention weights predicted by {\net}. Overall speaking, the frames with high attention weights do have clear image appearances for diagnosis. For example, the first three frames in Fig.~\ref{fig:vis}(b) clearly demonstrate the edge micro-lobulation and irregular shapes, which lead to malignant judgment. Furthermore, we plot the relationships between the predicted attention values and the feature distances to the centers. As shown in Fig.~\ref{fig:vis}(e), these two variables are linearly related, which indicates that {\net} the attention weights are effectively guided by the feature distances. 

The qualitative analysis proves the interpretability of our method, which will benefit clinical usage. Moreover, the attention weights reveal the importance of each frame for lesion diagnosis. Therefore, it can provide a new perspective for the keyframe extraction task of ultrasound videos.

\section{Conclusion}

We propose {\net}, a novel video classification model for breast ultrasound diagnosis. Our {\net} takes as input both the video data and image data to train the network. We propose the coherence loss to guide the training of the video model by the guidance of feature centers of the images. Our method significantly exceeds the performance of other competitive video baselines. The visualization of the attention weights validates the effectiveness and interpretability of our {\net}.


%
%
%
%
\clearpage

\end{document}